\documentclass{article} 
\usepackage{iclr2016_conference,times}
\usepackage{hyperref}
\usepackage{url}
\usepackage[pdftex]{graphicx}
\graphicspath{ {./figures/} }
\usepackage{natbib}
\usepackage{amsmath}
\usepackage{capt-of}
\usepackage[percent]{overpic}
\usepackage{textcomp}

\title{Unsupervised Learning of Visual Structure using Predictive Generative Networks}

\author{William Lotter, Gabriel Kreiman \& David Cox \\
Harvard University\\
Cambridge, MA 02138, USA \\
\texttt{\{lotter,davidcox\}@fas.harvard.edu} \\
\texttt{gabriel.kreiman@tch.harvard.edu} \\
}

\begin{document}

\maketitle

\begin{abstract}
The ability to predict future states of the environment is a central pillar of intelligence. At its core, effective prediction requires an internal model of the world and an understanding of the rules by which the world changes. 
Here, we explore the internal models developed by deep neural networks trained using a loss based on predicting future frames in synthetic video sequences, using a CNN-LSTM-deCNN framework.
We first show that this architecture can achieve excellent performance in visual sequence prediction tasks, including state-of-the-art performance in a standard ``bouncing balls'' dataset \citep{NIPS2008_3567}.
Using a weighted mean-squared error and adversarial loss \citep{NIPS2014_5423}, the same architecture successfully extrapolates out-of-the-plane rotations of computer-generated faces. 
Furthermore, despite being trained end-to-end to predict only pixel-level information, our Predictive Generative Networks learn a representation of the latent structure of the underlying three-dimensional objects themselves. 
Importantly, we find that this representation is naturally tolerant to object transformations, and generalizes well to new tasks, such as classification of static images. Similar models trained solely with a reconstruction loss fail to generalize as effectively. 
We argue that prediction can serve as a powerful unsupervised loss for learning rich internal representations of high-level object features.
\end{abstract}

\section{Introduction}
There is a rich literature in neuroscience concerning ``predictive coding,'' the idea that neuronal systems predict future states of the world and primarily encode only deviations from those predictions \citep{Rao1999, Summerfield2006, 10.3389/fpsyg.2012.00548, NIPS1999_1783, PCN2013, DBLP:journals/corr/ZhaoZWL14}.
While predicting future sensory inputs can be intrinsically useful, here we explore the idea that prediction might also serve as a powerful framework for unsupervised learning.
Not only do prediction errors provide a source of constant feedback, but successful prediction fundamentally requires a robust internal model of the world.

The problem of prediction is one of estimating a conditional distribution:  given recent data, estimate the probability of future states. There has been much recent success in this domain within natural language processing (NLP) \citep{DBLP:journals/corr/Graves13, NIPS2014_5346} and relatively low-dimensional, real-valued problems such as motion capture \citep{Frag2015,Gan2015}. 
Generating realistic samples for high dimensional images, particularly predicting the next frames in videos, has proven to be much more difficult. Recently, \citet{Ranzato2014} used a close analogy to NLP models by discretizing image patches into a dictionary set, for which prediction is posed as predicting the index within this set at future time points. 
This approach was chosen because of the innate difficulty of using traditional losses in video prediction. In pixel space, loss functions such as mean-squared error (MSE) are unstable to slight deformations and fail to capture intuitions of image similarity.
As illustrated by \citet{Sriv2015}, predictive models trained with MSE tend to react to uncertainty with blurring. \citet{Sriv2015} propose a Long Short-Term Memory (LSTM) \citep{hoch} Encoder-Decoder model for next frame prediction and, despite blurry predictions on natural image patches, their results point to the potential of prediction as unsupervised learning, as predictive pre-training improved performance on two action recognition datasets.

A promising alternative to MSE is an adversarial loss, as in the Generative Adversarial Network (GAN) \citep{NIPS2014_5423}. This framework involves training a generator and discriminator in a minimax fashion. Successful extensions, including a conditional GAN \citep{Gauthier2014, MirzaO14} and a Laplacian pyramid of GANs \citep{Denton15}, show its promise as a useful model for generating images. 

Here we build upon recent advances in generative models, as well as classical ideas of predictive coding and unsupervised temporal learning, to investigate deep neural networks trained with a predictive loss. We use a model consisting of a convolutional neural network (CNN), an LSTM, and a deconvolutional neural network (deCNN) \citep{invertCNN, tejas}. Falling in the class of Encoder-Recurrent-Decoder (ERD) architectures \citep{Frag2015}, our model is trained end-to-end to combine feature representation learning with the learning of temporal dynamics. 
In addition to MSE, we implement an adversarial loss (AL). 

We demonstrate the effectiveness of our architecture, trained with MSE, on a standard ``bouncing balls'' experiment \citep{NIPS2008_3567} before applying the same architecture to a dataset of computer-generated faces undergoing rotations. This dataset is an appropriate intermediate step between toy examples and full-scale natural images, where we can more fully study the representational learning process.
We find that a weighted combination of MSE and AL leads to predictions that are simultaneously consistent with previous frames and visually convincing. 
Furthermore, over the course of training, the model becomes better at representing the latent variables of the underlying generative model.
We test the generality of this representation in a face identification task requiring transformation tolerance.
In the classification of \textit{static} images, the model, trained with a predictive loss on dynamic stimuli, strongly outperforms comparable models trained with a reconstruction loss on static images.
Thus, we illustrate the promise of prediction as unsupervised model learning from video.

\section{Related Work}
\label{related_work}

In addition to the work already cited, the current proposal has strong roots in the idea of learning from temporal continuity. Early efforts in this field demonstrated how invariances to particular transformations can be learned through temporal exposure \citep{fold}. Related algorithms, such as Slow Feature Analysis (SFA) \citep{sfa}, take advantage of the persistence of latent causes in the world to learn representations that are robust to noisy, quickly-varying sensory input. More recent work has explicitly implemented temporal coherence in the cost function of deep learning architectures, enforcing the networks to develop a representation where feature vectors of consecutive video frames are closer together than those between non-consecutive frames \citep{mohabi,goroshin,wang}. Related to these ideas, a recent paper proposed training models to linearize transformations observed over sequences of frames in natural video \citep{goroshin2}.

Also related to our approach, especially in the context of rotating objects, is the field of relational feature learning \citep{mem2007, taylor2009}. This posits modeling time-series data as learning representations of the \textit{transformations} that take one frame to the next. Recently, \citet{NIPS2014_5549} proposed a predictive training scheme where a transformation is first inferred between two frames and is then applied again to obtain a prediction of a third frame. They reported evidence of a benefit of using predictive training versus traditional reconstruction training. 


Finally, using variations of autoencoders for unsupervised learning and pre-training, is certainly not new \citep{erhan,ben2006}. In fact, \citet{Palm2012} coined the term ``Predictive Encoder'' to refer to an autoencoder that is trained to predict future input instead of reconstructing current stimuli. In preliminary experiments, it was shown that such a model could learn Gabor-like filters in training scenarios where traditional autoencoders failed to learn useful representations. 


\section{Predictive Generative Networks}
\label{pgn}

\begin{figure}[h]
\begin{center}
\includegraphics[width=0.65\textwidth]{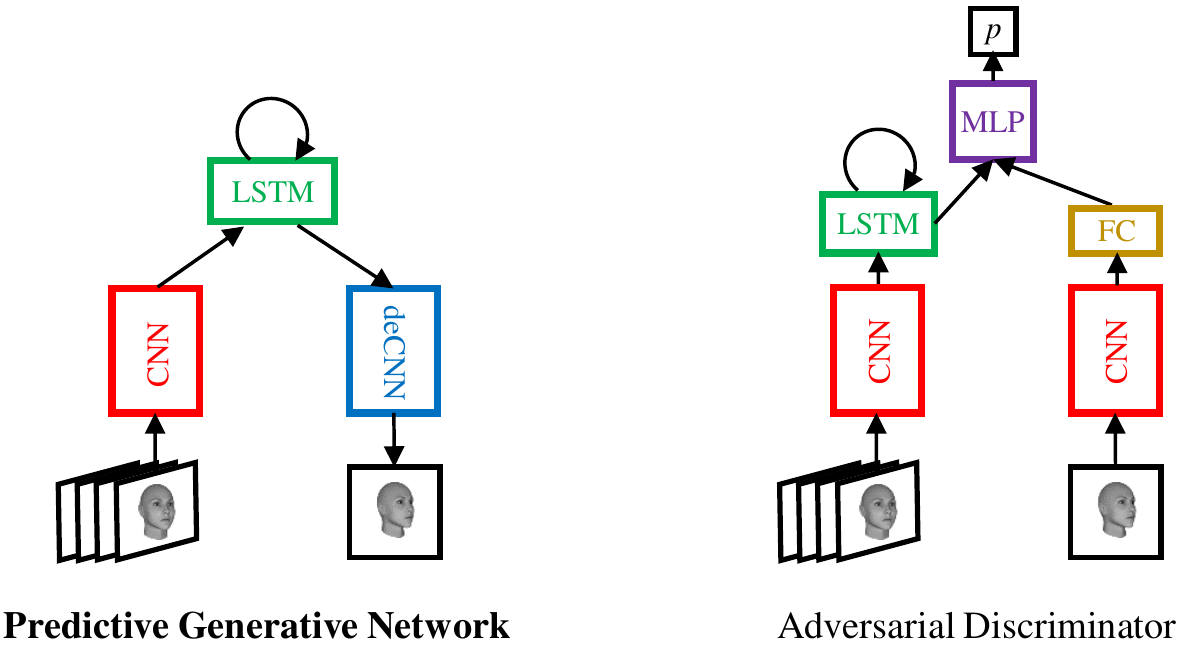}

\end{center}
\caption{Predictive Generative Network (PGN)}
\label{fig1}
\end{figure}

A schematic of our framework is shown in Figure~\ref{fig1}. The generative model first embeds a sequence of frames, successively, into a lower-dimensional feature space using a CNN. We use a CNN consisting of two layers of convolution, rectification, and max-pooling. The CNN output is passed to an LSTM network \citep{hoch}. Briefly, LSTM units are a particular type of hidden unit that improve upon the vanishing gradient problem that is common when training RNNs \citep{ben1994}. An LSTM unit contains a cell, $c_t$, which can be thought of as a memory state. Access to the cell is controlled through an input gate, $i_t$, and a forget gate, $f_t$. The final output of the LSTM unit, $h_t$, is a function of the cell state, $c_t$, and an output gate, $o_t$. We use a version of the LSTM with the following update equations:
\begin{align*}
i_t &= \sigma ( W_{xi}x_t + W_{hi}h_{t-1}+b_i ) \\
f_t &= \sigma (W_{xf}x_t + W_{hf}h_{t-1}+b_f) \\
c_t &= f_t c_{t-1}+i_t \tanh (W_{xc} x_t + W_{hc} h_{t-1} + b_c) \\
o_t &= \sigma (W_{xo} x_t + W_{ho} h_{t-1} + b_o) \\
h_t &= o_t \tanh (c_t)
\end{align*}

where $x_t$ is the input to the LSTM network, $W_{\bullet \bullet}$ are the weight matrices, and $\sigma$ is the elementwise logistic sigmoid function. We use $1568$ LSTM units for the bouncing balls dataset and $1024$ for the rotating faces. All models were implemented using the software package Keras \citep{keras}.

Upon processing the last output of the CNN, the LSTM hidden state is outputted to a deCNN, which produces a predicted image. 
The parameters of the network are chosen such that the predicted image is the same size of the input. For the rotating faces dataset, the deCNN consists of a fully-connected (FC) layer, followed by two layers of nearest-neighbor upsampling, convolution, and rectification. The last layer also contains a saturating non-linearity set at the maximum pixel value. Due to the lower dimensional size, the FC layer of the deCNN was omitted for the model trained on the bouncing balls dataset.

With adversarial loss, we also considered concatenating the LSTM hidden state with a random vector before it is passed to the deCNN, which would allow sampling as in previous conditional GANs \citep{Gauthier2014, MirzaO14}. 
However, in our case, the sampling ended up being highly peaked, with different samples from the same input sequence being nearly indistinguishable, possibly because our sequences are deterministic, so the random vector was later discarded. 

For adversarial loss, the predicted frame from the generator is passed to a discriminator network, which is also conditioned on the original input sequence. Similar to the generator, the input sequence is passed through a CNN and LSTM.  We experimented with sharing the CNN and LSTM weights between the discriminator and generator, but ultimately had best results with each network having its own separate set of parameters. After processing the last input frame, the LSTM hidden state of the discriminator is concatenated with an encoding of the proposed next frame in the sequence, either the true frame of the generator's output. The encoding consists of a CNN and a FC layer, resulting in feature vector with the same size of the LSTM output. The concatenation of these two vectors is passed to a multi-layer perceptron (MLP), consisting of three FC layers with a sigmoidal read-out. 

We use the original formulation of the adversarial loss function \citep{NIPS2014_5423}. The discriminator outputs a probability that a proposed frame came from the ground truth data. It is trained to maximize this probability when the frame came from the true distribution and minimize it when it is produced by the generator. The generator is trained to fool the discriminator. Let $x_{1:t}^i$ be an input sequence of $t$ frames and $x_{t+1}^i$ be the true next frame. Let the proposed frame from the generator be $G(x_{1:t}^i)$ and $D(\bullet,x_{1:t}^i)$ be the discriminator's output. Given a mini-batch size of $n$ sequences, the loss of the discriminator, $L_{D}^{(AL)}$, and of the generator, $L_{G}^{(AL)}$, have the form:
\begin{align*}
L_{D}^{(AL)} &= -\frac{1}{2n}\sum_{i=1}^n \lbrack \log D(x_{t+1}^i,x_{1:t}^i)  + \log(1-D(G(x_{1:t}^i),x_{1:t}^i)) \rbrack \\
L_{G}^{(AL)} &= \frac{1}{n}\sum_{i=1}^n \log(1-D(G(x_{1:t}^i),x_{1:t}^i)) 
\end{align*}

As in the original paper \citep{NIPS2014_5423}, we actually train the generator to maximize $\log(D(G(x_{1:t}^i),x_{1:t}^i))$ and not minimize $\log(1-D(G(x_{1:t}^i),x_{1:t}^i))$, as the latter tends to saturate early in training. 

Due to complementary roles for MSE and AL in training the generator, we combine both in a weighted loss, $L_G^{(tot)}$, controlled by a hyperparameter $\lambda$:
\begin{align*}
L_G^{(tot)} &= L_G^{(MSE)} + \lambda L_G^{(AL)}
\end{align*}
Even for high values of $\lambda$, the MSE loss proved to be useful as it stabilizes and accelerates training. 

\section{Prediction Performance}

We evaluated the Predictive Generative Networks (PGNs) on two datasets of synthetic video sequences. As a baseline to compare against other architectures, we first report performance on a standard bouncing balls paradigm \citep{NIPS2008_3567}. We then proceed to a dataset containing out-of-the-plane rotations of computer-generated faces, where we thoroughly analyze the learned representations.




\subsection{Bouncing Balls}

\begin{figure}[t]
	\centering
	\begin{minipage}[h]{0.3\textwidth}
		
		\centering
		\vspace{0pt}
		\captionof{table}{Average prediction error for the bouncing balls dataset. $^\dagger$\citep{Gan2015} $^\diamond$\citep{Mittelman2014}} 
		\begin{tabular}{ll}

			\multicolumn{1}{c}{\bf Model}  &\multicolumn{1}{c}{\bf Error}
			\\ \hline \\
			\bf PGN (MSE)        & \boldmath$0.65 \pm 0.11$ \\
			DTSBN $^\dagger$            &$2.79 \pm 0.39$ \\
			SRTRBM $^\diamond$            &$3.31 \pm 0.33$ \\
			RTRBM $^\diamond$            &$3.88 \pm 0.33$ \\
			Frame $t$-1            &$11.86 \pm 0.27$ \\
		\end{tabular}
		\label{table1}
	\end{minipage}\hfill
	\begin{minipage}[h]{0.6\textwidth}
		\centering
		\vspace{0pt}
		\includegraphics[width=\textwidth]{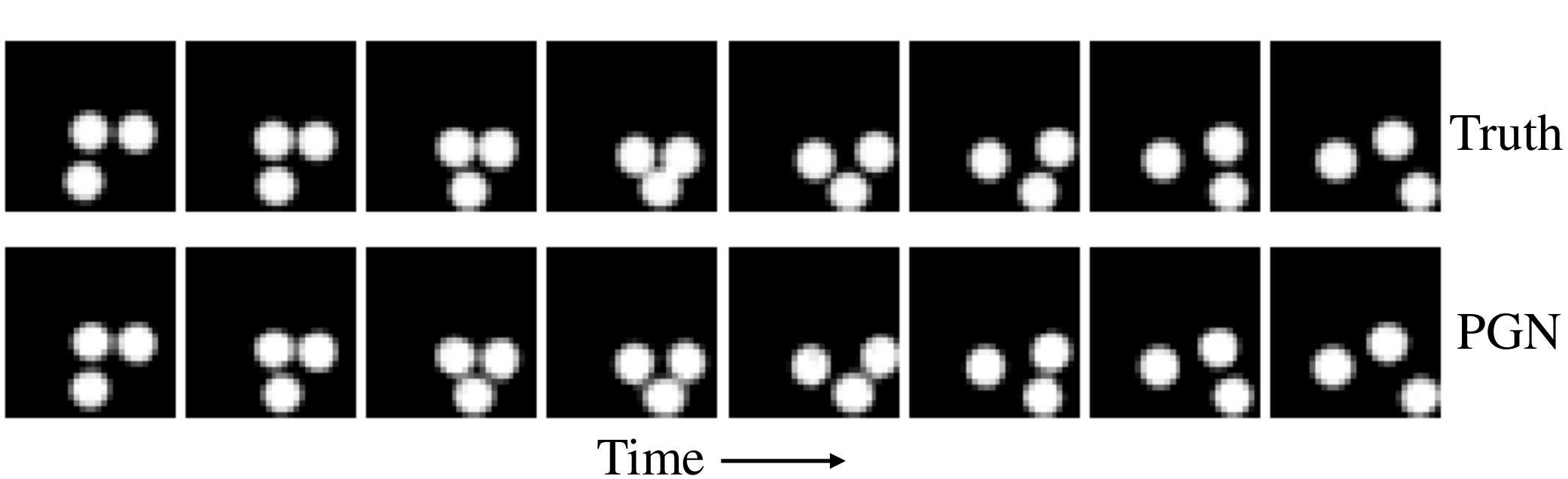}
		
		\caption{Example prediction sequence for bouncing balls dataset. Predictions are repeatedly generated one step ahead using the prior ten frames as input.}
		\label{fig2}
	\end{minipage}
\end{figure}

The bouncing balls dataset is a common test set for models that generate high dimensional sequences. It consists of simulations of three balls bouncing in a box. We followed standard procedure to create $4000$ training videos and $200$ testing videos \citep{NIPS2008_3567} and used an additional $200$ videos for validation. Our networks were trained to take a variable number of frames as input, selected randomly each epoch from a range of $5$ to $15$, and output a prediction for the next timestep. Training with MSE was very effective for this dataset, so AL was not used. Models were optimized using RMSprop \citep{rms} with a learning rate of $0.001$. In Table~\ref{table1}, we report the average squared one-step-ahead prediction error per frame. Our model compares favorably to the recently introduced Deep Temporal Sigmoid Belief Network \citep{Gan2015} and restricted Boltzmann machine (RBM) variants, the recurrent temporal RBM (RTRBM) and the structured RTRBM (SRTRBM) \citep{Mittelman2014}. An example prediction sequence is shown in Figure~\ref{fig2}, where each prediction is made one step ahead by using the ten previous frames as input.

\subsection{Rotating Faces}

For the rotating faces dataset, each video consists of a unique, randomly generated face rotating about the vertical axis with a random speed and initial angle. Speed is sampled uniformly from $[0,\pi/6]$ rad/frame with an initial angle sampled from $[-\pi/2,\pi/2]$, where $0$ corresponds to a frontal view. Input sequences consist of $5$ frames of size $150$x$150$ pixels.  We use $4000$ clips for training and $200$ for validation and testing. 

Generative models are often evaluated using a Parzen window estimate of the log-likelihood \citep{parzen}, but due to the deficiencies of this approach for high dimensional images, we chose values of the weighting parameter between MSE and AL, $\lambda$, based on qualitative assessment.
Adversarial models are notoriously difficult to train \citep{eyescream} and we empirically found benefits in giving the generator and discriminator a ``warm start''.
For the generator, this corresponded to initializing from a solution trained solely with MSE. 
This is analogous to increasing the value of $\lambda$ over training, thus ensuring that the models learn the low frequency components first, and then the high frequency components, akin to the LAPGAN approach \citep{Denton15}.
For the discriminator, we used a pre-training scheme where it was first trained against a generator with a high value of $\lambda$.
This exposes the discriminator to a wide variety of stimuli in pixel space early in training, which helps it quickly discriminate between real and generated images when it is subsequently paired with the MSE-initialized generator.
For the data shown in this paper, these initialization schemes are used and $\lambda$ is set to $0.0002$.
The generator is optimized using RMSprop \citep{rms} with a learning rate of $0.001$. The discriminator is trained with stochastic gradient descent (SGD), with a learning rate of $0.01$ and momentum of $0.5$.

\begin{figure}[h]
\begin{center}
\includegraphics[width=\textwidth]{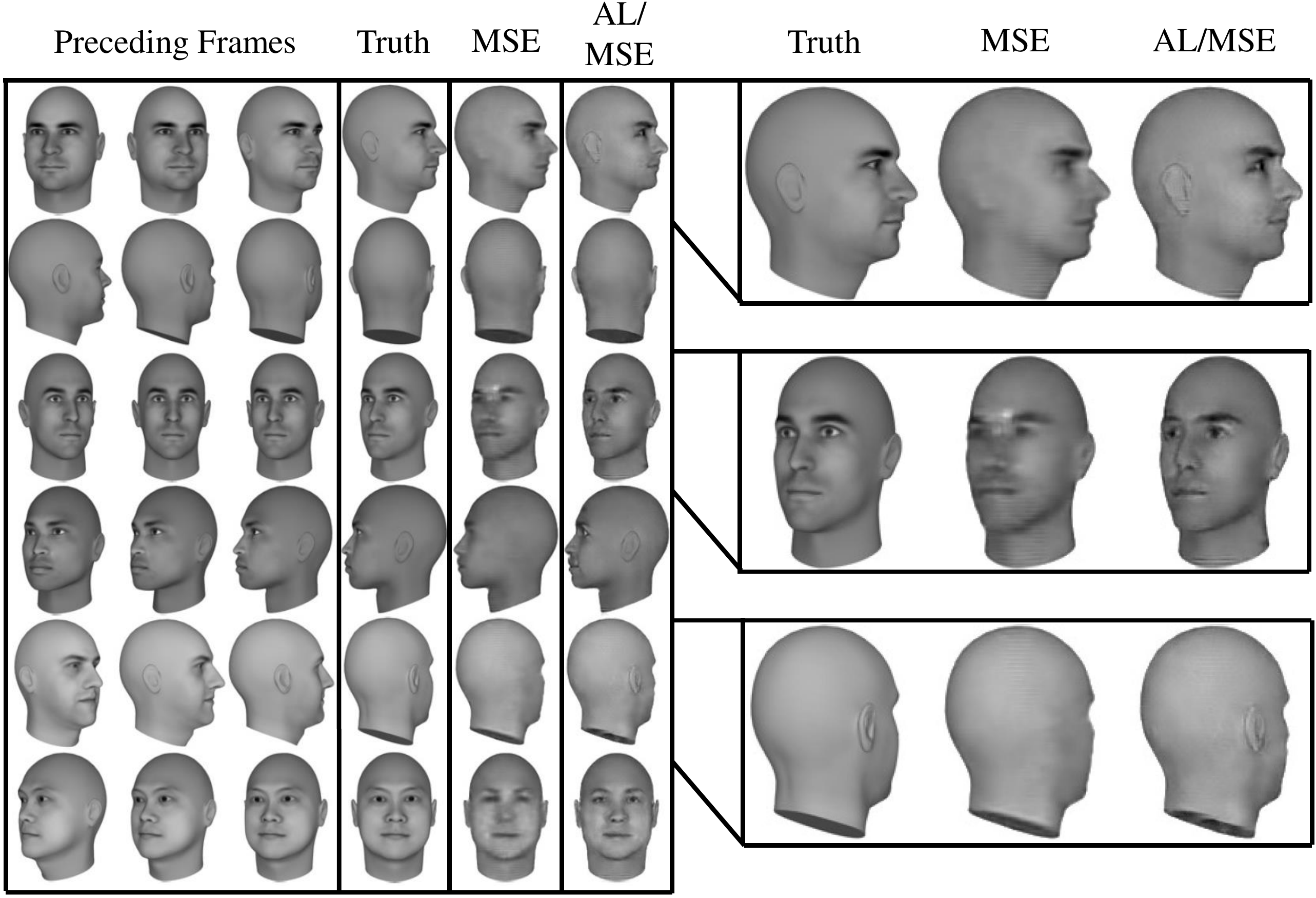}
\end{center}
\caption{Example predictions for the rotating faces dataset. Predictions for models trained with MSE and a weighted MSE and adversarial loss (AL) are shown.}
\label{fig3}
\end{figure}

Example predictions are shown in Figure~\ref{fig3}. We compare the results of training with MSE to the weighted AL/MSE model. Predictions are for faces not seen during training. Both models successfully estimate the angle and basic shape of the face very well. However, the MSE model produces blurred, low-passed versions as expected, whereas the AL/MSE model generates images with high detail. Most notably, the AL/MSE model has learned that faces contain conspicuous eyes and ears, which are largely omitted by the MSE model. 
When the AL/MSE model does make mistakes, it's often through generating faces that notably look realistic, but seem slightly inconsistent with the identity of the face in the preceding frames. This can be seen in the second row in the right panel of Figure~\ref{fig3}. Weighting AL higher exaggerates this effect. One would hope that the discriminator would be able to discern if the identity changed for the proposed rotated view, but interestingly, even humans struggle with this task \citep{wallis}.

\section{Exploring Latent Representation Learning}

Beyond generating realistic predictions, we are interested in understanding the representations learned by the predictive models, especially in relation to the underlying generative model. The faces are created according to a principal component analysis (PCA) in ``face space'', which was estimated from real-world faces. In addition to the principal components (PCs), the remaining latent variables are the initial angle and rotation speed. 

A decoding analysis was performed in which an L2-regularized regression was used to estimate the latent variables from the LSTM representation. We decoded from the hidden unit responses after five time steps, i.e. the last time step before the hidden representation is outputted to the deCNN to produce a predicted image. The regression was fit, validated, and tested using a different dataset than the one used to train the model. 

As a baseline, we compare decoding from the predictive models to a model with the same architecture, trained on precisely the same stimulus set, but with a reconstruction loss. 
Here, the input sequence is all six frames, for which the model is trained to reconstruct the last. 
Note, the model cannot simply copy the input, but must learn a low dimensional representation, because the LSTM has a dimension size much less than the input ($1024$ v.s. $150$x$150=22.5$K), e.g. the common autoencoder scenario.

The decoding results for the initial angle, rotation speed, and first four principal components are contained in Table~\ref{table2}.
Although they produce visually distinct predictions, the MSE and AL/MSE PGNs show similar decoding performance.
This is not surprising since the PCs dictate the shape of the face, which the MSE model estimates very well.
Nevertheless, both predictive models strongly outperform the autoencoder.  
There are more sophisticated ways to train autoencoders, including denoising criteria \citep{DAE}, but here we show that, for a fixed training set, a predictive loss can lead to a better representation of the underlying generative model than a reconstruction loss. 

\begin{table}[h]
\caption{Decoding accuracy ($r^2$) of latent variables from the LSTM hidden unit representation.}
\begin{center}
\begin{tabular}{ |c|c|c|c|c|c|c| } 
\hline
\bf Model  &\bf Angle &\bf Speed &\bf PC1 &\bf PC2 &\bf PC3 &\bf PC4 \\
\hline
PGN (MSE)         &$0.994$ &$0.986$ &$0.877$ &$0.826$ &$0.723$ &$0.705$ \\ 
PGN (AL/MSE)             &$0.994$ &$0.990$ &$0.873$ &$0.828$ &$0.724$ &$0.686$ \\
Autoencoder (MSE)             &$0.943$ &$0.927$ &$0.834$ &$0.772$ &$0.655$ &$0.635$\\
\hline
\end{tabular}
\end{center}
\label{table2}
\end{table}

\begin{figure}[h]
\begin{center}
\includegraphics[width=0.8\textwidth]{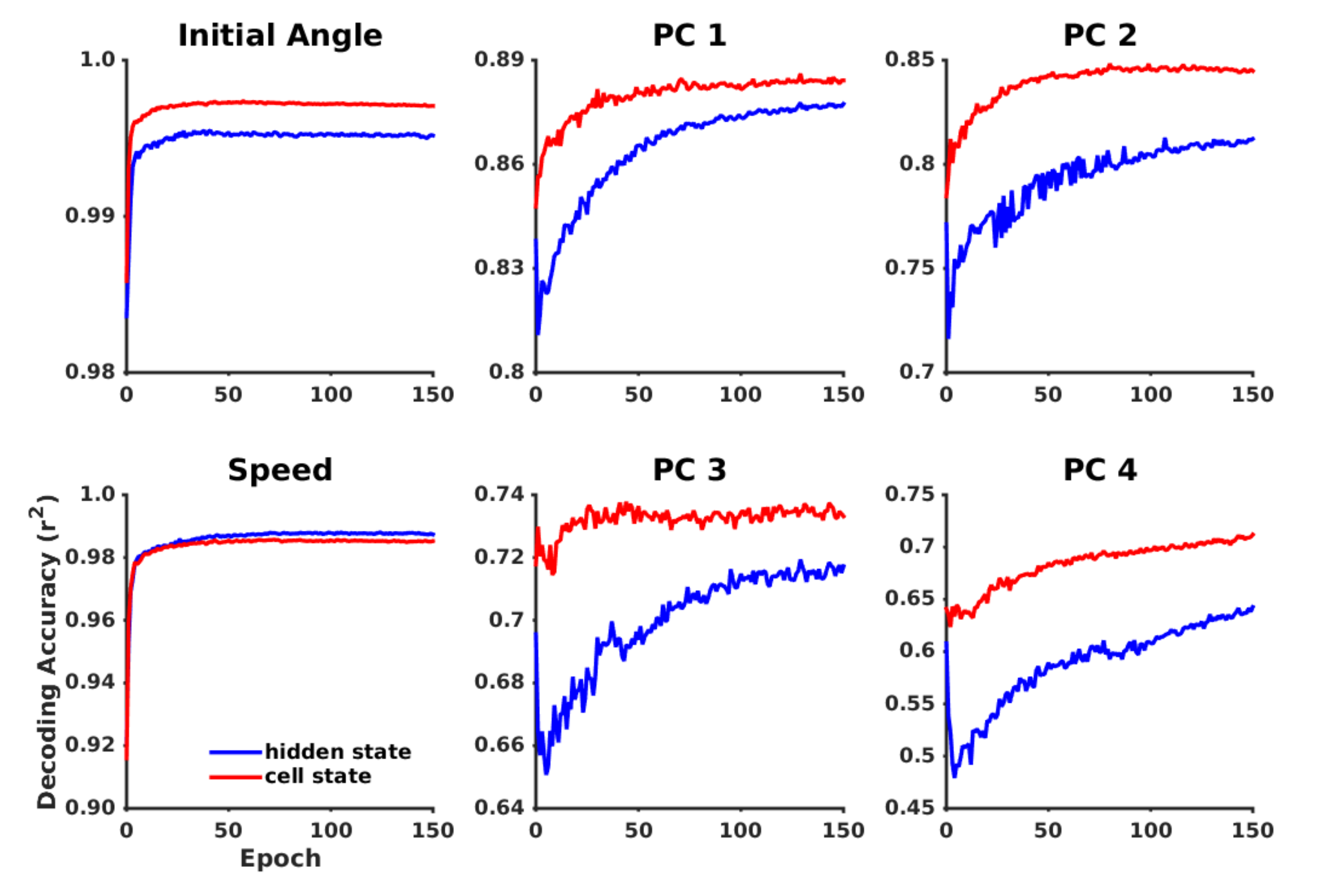}
\end{center}
\caption{Dynamics of latent variable decoding from internal representation of PGN (MSE)}
\label{fig4}
\end{figure}

To gain insight into the learning dynamics, we show decoding performance for both the hidden state and cell state as a function of training epoch for the MSE model in Figure~\ref{fig4}. Epoch $0$ corresponds to the random initial weights, from which the latent variables can already be decoded fairly well, which is expected given the empirical evidence and theoretical justifications for the success of random weights in neural networks \citep{jarrett,pinto2009,saxe_random}.
Still, it is clear that the ability to estimate all latent variables increases over training. The model quickly peaks at its ability to linearly encode for speed and initial angle. 
The PCs are learned more slowly, with decoding accuracy for some PCs actually first decreasing while speed and angle are rapidly learned. The sequence in which the model learns is reminiscent of theoretical work supporting the notion that modes in the dataset are learned in a coarse-to-fine fashion \citep{Saxe_learninghierarchical}.

\begin{figure}[h]
	\centering
	\begin{minipage}[h]{0.30\textwidth}
		\centering
		\vspace{0pt}
		\includegraphics[width=\textwidth]{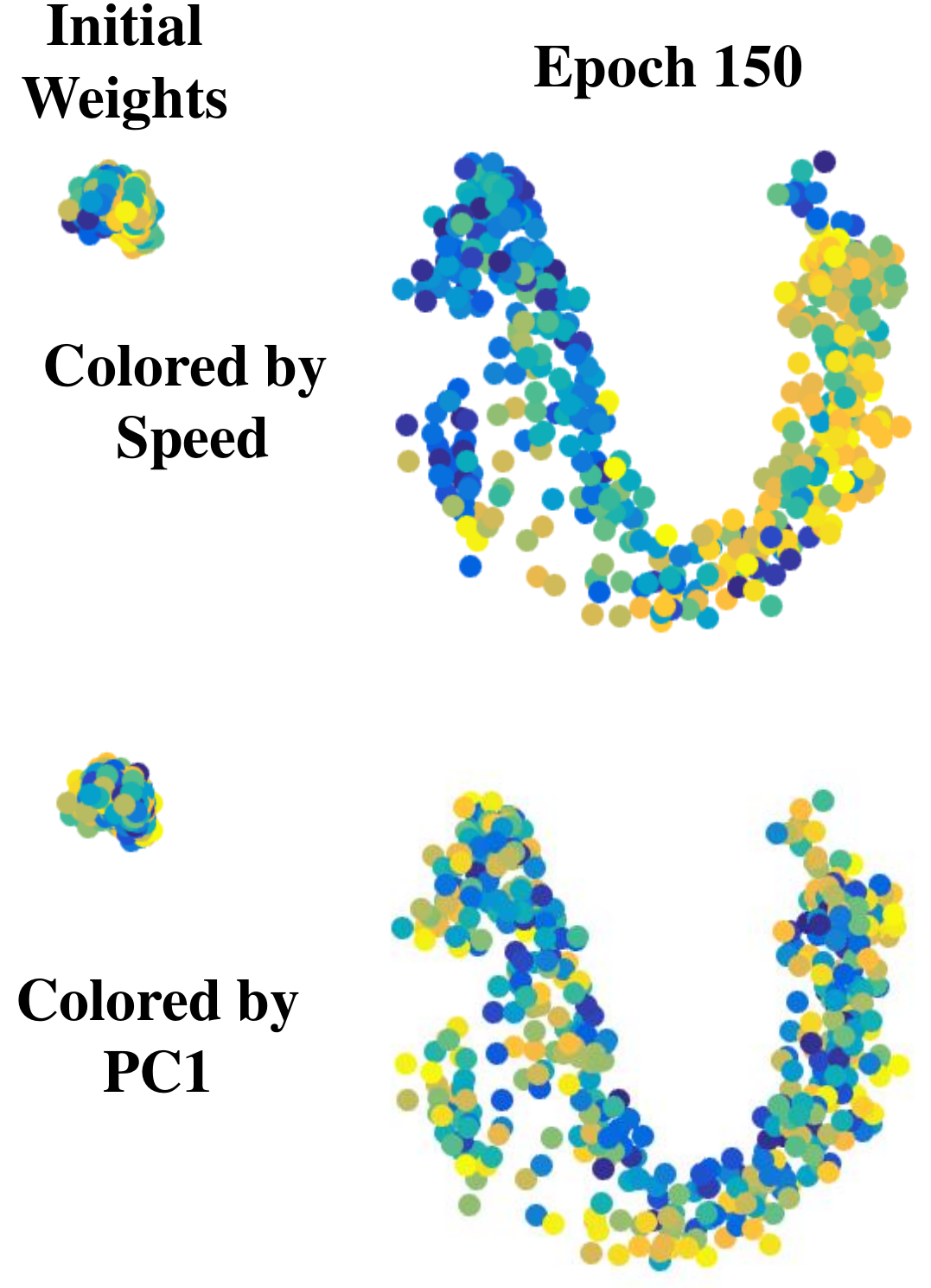}
		
		\caption{Multidimensional scaling plot of the LSTM representation demonstrating changes with training.}
		\label{fig5}
	\end{minipage}\hfill
	\begin{minipage}[h]{0.62\textwidth}
		\centering
		\vspace{0pt}
		\includegraphics[width=\textwidth]{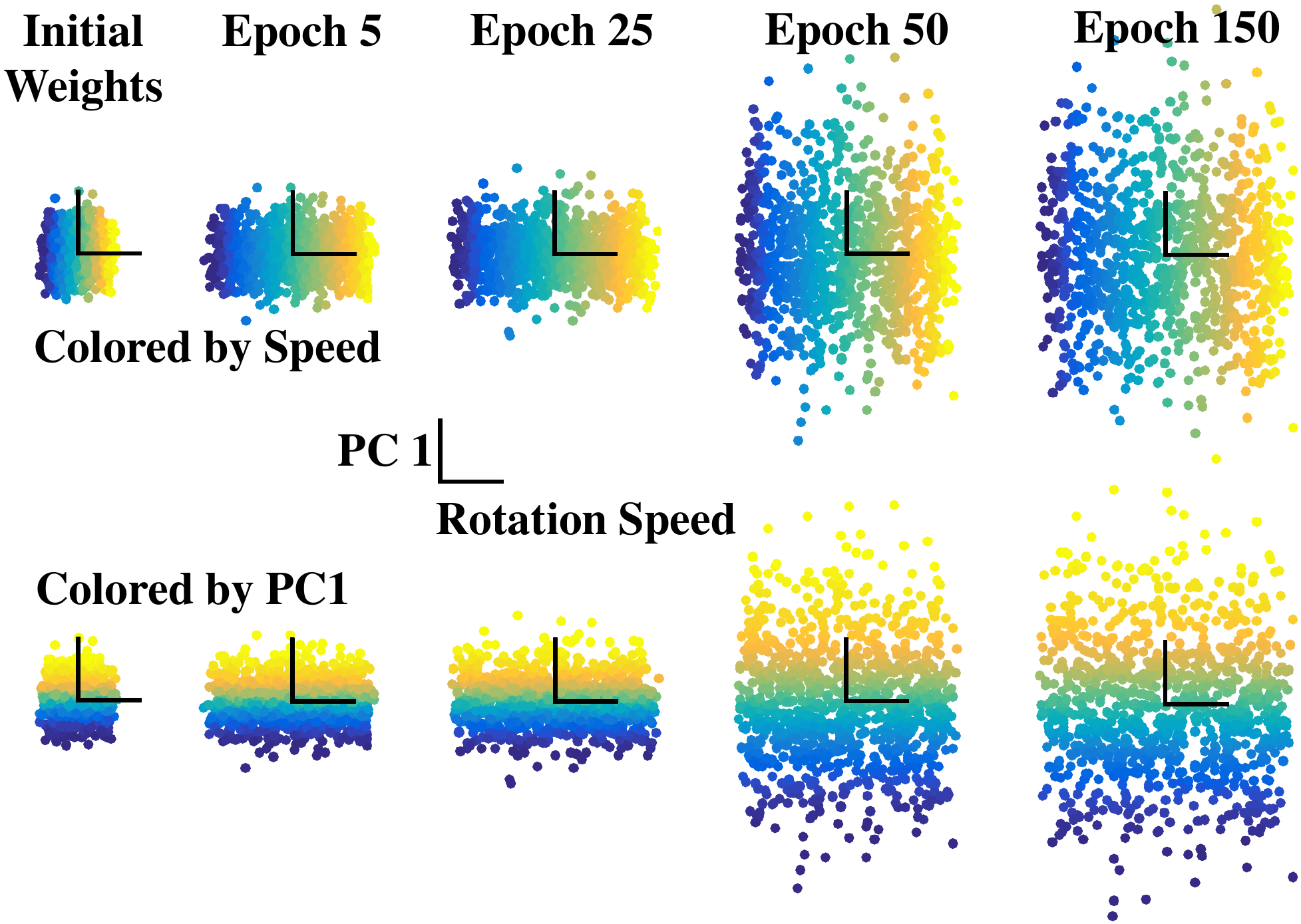}
		
		\caption{Projection of LSTM feature space on latent variables axes. Axes are in the direction of regression coefficients. A different regression was fit for each epoch.}
		\label{fig6}
	\end{minipage}
\end{figure}

To understand the representational changes accompanying the increase in decoding performance, we provide visualizations of the hidden unit feature space over training in Figures~\ref{fig5} and ~\ref{fig6}. Figure~\ref{fig5} contains a multidimensional-scaling (MDS) plot for the initial random weights and the weights after Epoch $150$ trained with MSE. Points are colored by PC1 value and rotation speed. Although a regression on this feature space at Epoch $0$ produces an $r^2$ of \texttildelow $0.83$, it is apparent that the structure of this space changes with training. To have a more clear understanding of these changes, we linearized the feature space in two dimensions with axes pointing in the direction of the regression coefficients for decoding PC$1$ and rotation speed. Points from a held-out set were projected on these axes and plotted in Figure~\ref{fig6} and we show the evolution of the projection space, with regression coefficients calculated separately for each epoch. Over training, the points become more spread out over this manifold. This is not due to an overall increase in feature vector length, as this does not increase over training. Thus, with training, the variance in the feature vectors become more aligned with the latent variables of the generative model. 

\begin{figure}[h]
\begin{center}
\includegraphics[width=\textwidth]{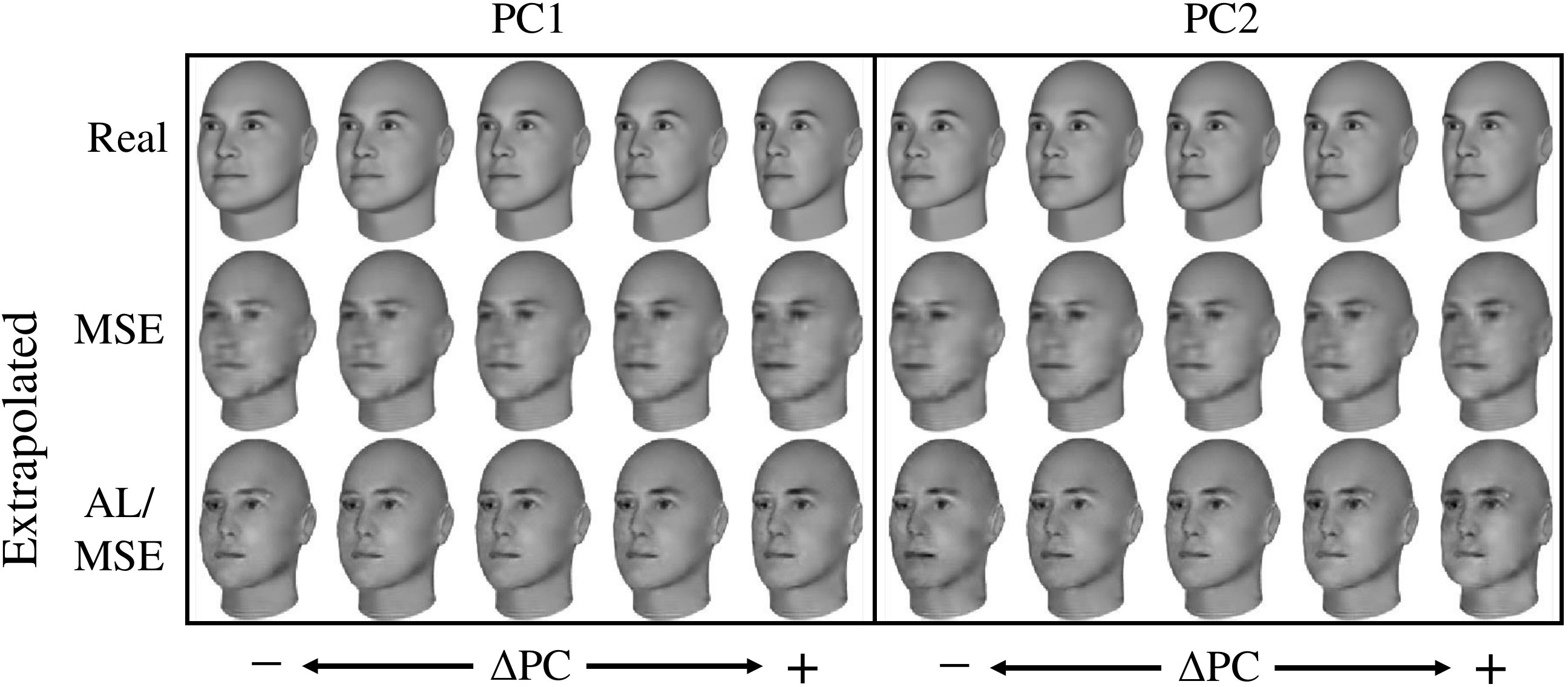}

\end{center}
\caption{Linearly moving through LSTM feature space along principal component axes.}
\label{fig7}
\end{figure}

The previous analyses suggest that the PGNs learn a low dimensional, linear representation of the face space. This is further illustrated in Figure~\ref{fig7}. Here, the feature representation of a given seed face is calculated and then the feature vector is perturbed in the direction of a principal component axis, as estimated in the decoding analysis.
The new feature vector is then passed to the pre-trained deCNN to produce an image. The generated image is compared with changing the PC value directly in the face generation software. Figure~\ref{fig7} shows that the extrapolations produce realistic images, especially for the AL/MSE model, which correlate with the underlying model. 
The PC dimensions do not precisely have semantic meanings, but differences can especially be noticed in the cheeks and jaw lines. The linear extrapolations in feature space generally match changes in these features, demonstrating that the models have learned a representation where the latent variables are linear.

\begin{figure}[h]
\begin{center}
\includegraphics[width=0.6\textwidth]{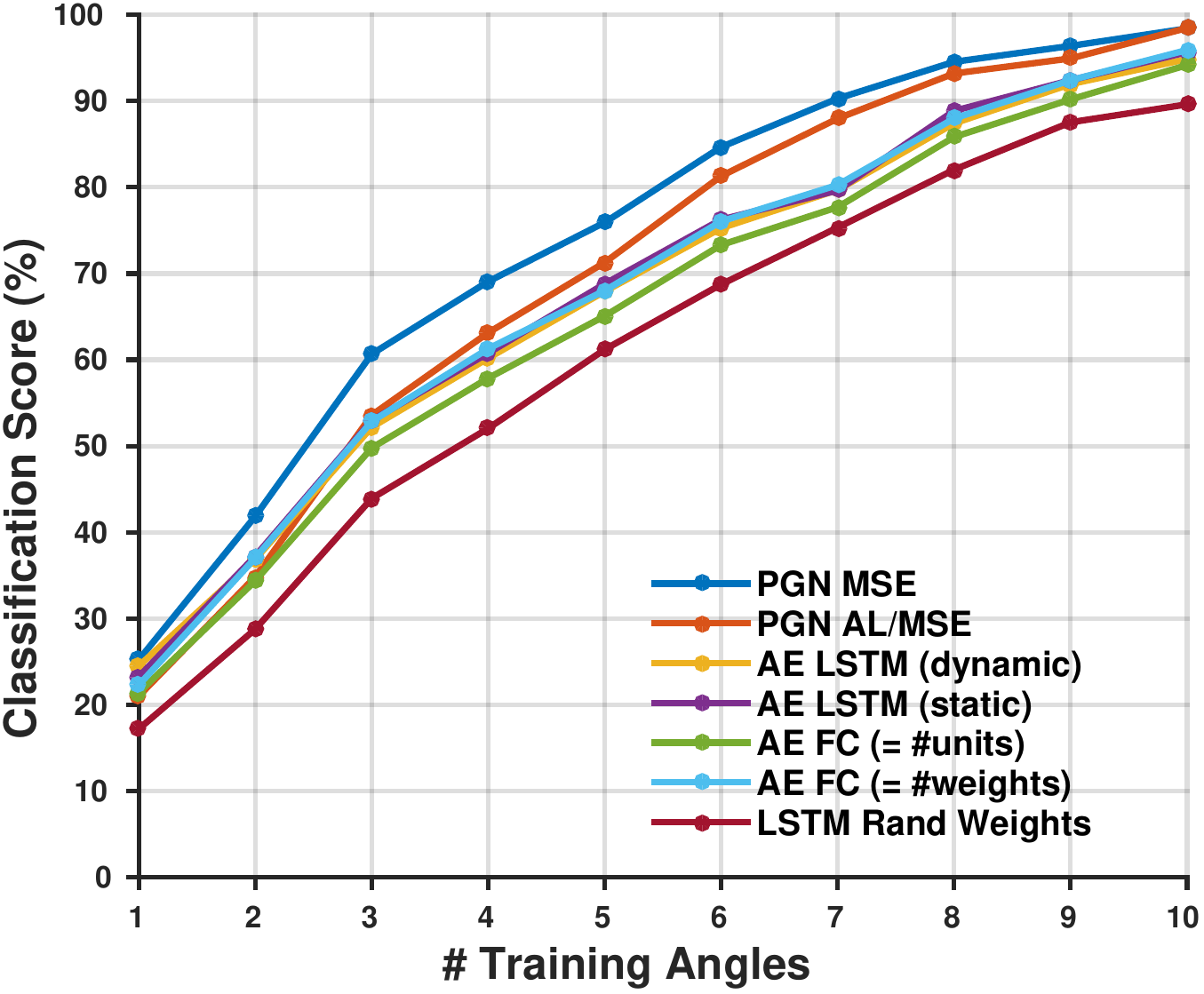}

\end{center}
\caption{Classification accuracy on a $50$-way face identification task. AE: Autoencoder.}
\label{fig8}
\end{figure}

While the generation of frame-by-frame future predictions is useful \emph{per se}, we were especially interested in the extent to which prediction could be used as an unsupervised loss for learning representations that are suited to other tasks. 
We tested this hypothesis through a task completely orthogonal to the original loss function, namely classification of static images. As a control, to specifically isolate the effect of the loss itself, we trained comparable models using a reconstruction loss and either dynamic or static stimuli. The first control was carried over from the latent variable decoding analysis and had the same architecture and training set of the PGN, but was trained with a reconstruction loss (denoted as \textbf{AE LSTM (dynamic)} in Fig.~\ref{fig8}). The next model again had the same architecture and a reconstruction loss, but was trained on static videos [\textbf{AE LSTM (static)}]. A video was created for each unique frame in the original training set. 
For the last two models, the LSTM was replaced by a fully-connected (FC) layer, one with an equal number of weights [\textbf{AE FC (= \# weights)}] and the other with an equal number of units [\textbf{AE FC (= \# units)}] as the LSTM. These were trained in a more traditional autoencoder fashion to simply reconstruct single frames, using every frame in the original video set. All control models were trained with MSE since AL is more sensitive to hyperparameters.

The classification dataset consisted of $50$ randomly generated faces at $12$ equally-spaced angles between $[-\frac{\pi}{2},\frac{\pi}{2}]$. A support vector machine (SVM) was fit on the feature representations of each model. For the models containing the LSTM layer, the feature representation at the fifth time step was chosen. To test for transformation tolerance, training and testing were done with separate sets of angles. 

The classification performance curves are shown in Figure~\ref{fig8}. While all models show improvement compared to random initial weights, the predictive models outperform the controls. The MSE PGN has the highest score for each size of the training data. The AL/MSE PGN performs moderately worse, but still better than the control models. This is likely because, as previously mentioned, the AL/MSE model can tend to produce realistic faces, but are somewhat unfaithful to the underlying identity.
While this is a relatively simple task compared to modern machine learning datasets, it provides a proof-of-principle that a model trained with a unsupervised, predictive loss on dynamic sequences can learn interesting structure, which is even useful for other tasks. 

\section{Conclusion}

In extending ideas of predictive coding and learning from temporal continuity to modern, deep learning architectures, we have shown that an unsupervised, predictive loss can result in a rich internal representation of visual objects. 
Our CNN-LSTM-deCNN models trained with such a loss function successfully learn to predict future image frames in several contexts, ranging from the physics of simulated bouncing balls to the out-of-plane rotations of previously unseen computer-generated faces. 
However, importantly, models trained with a predictive unsupervised loss are also well-suited for tasks beyond the domain of video sequences. For instance, representations trained with a predictive loss outperform other models of comparable complexity in a supervised classification problem with static images.
This effect is particularly pronounced in the regime where a classifier must operate from just a few example views of a new object (in this case, face).
Taken together, these results support the idea that prediction can serve as a powerful framework for developing transformation-tolerant object representations of the sort needed to support one- or few-shot learning.

The experiments presented here are all done in the context of highly-simplified artificial worlds, where the underlying generative model of the stimuli is known, and where the number of degrees of freedom in the data set are few.
While extending these experiments to real world imagery is an obvious future priority, we nonetheless argue that experiments with highly controlled stimuli hold the potential to yield powerful guiding insights.
Understanding how to scale predictive generative models of this form to encompass all of the transformation degrees of freedom found in real-world objects is an area of great interest for future research.

\subsubsection*{Acknowledgments}
We would like to thank Chuan-Yung Tsai for fruitful discussions.  This work was supported by a grant from the National Science Foundation (NSF IIS 1409097) and by the Center for Brains, Minds and Machines (CBMM), funded by NSF STC award CCF-1231216.

\bibliography{predictive_network_manuscript_iclr_2015}
\bibliographystyle{iclr2016_conference}

\end{document}